\documentclass[10pt, a4paper]{article}

\usepackage{lrec-coling2024} 

\title{JaParaPat: A Large-Scale Japanese-English Parallel Patent
  Application Corpus}

\name{Masaaki Nagata, Katsuki Chousa, Norihito Yasuda} 

\address{NTT Communication Science Laboratories, NTT Corporation \\
  2-4 Hikaridai Seika-cho Souraku-gun Kyoto-fu 619-0237 Japan \\
  \{masaaki.nagata,katsuki.chousa,norihito.yasuda\}@ntt.com\\}

\abstract{ We constructed JaParaPat (\underline{Ja}panese-English
  \underline{Para}llel \underline{Pat}ent Application Corpus), a
  bilingual corpus of more than 300 million Japanese-English sentence
  pairs from patent applications published in Japan and the United
  States from 2000 to 2021.  We obtained the publication of unexamined
  patent applications from the Japan Patent Office (JPO) and the
  United States Patent and Trademark Office (USPTO).  We also obtained
  patent family information from the DOCDB, that is a bibliographic
  database maintained by the European Patent Office (EPO).  We
  extracted approximately 1.4M Japanese-English document pairs, which
  are translations of each other based on the patent families, and
  extracted about 350M sentence pairs from the document pairs using a
  translation-based sentence alignment method whose initial
  translation model is bootstrapped from a dictionary-based sentence
  alignment method.  We experimentally improved the accuracy of the
  patent translations by 20 bleu points by adding more than 300M
  sentence pairs obtained from patent applications to 22M sentence
  pairs obtained from the web.
  \\
  \newline  }

\begin{document}

\maketitleabstract

\section{Introduction}

International patent applications are numerous but finite. In this
work, we aim to disclose the quantity and the quality of the parallel
data obtainable from international patent applications in Japanese and
English and the potential translation accuracy using these
resources. Since most translation for international patent
applications in Japan involves Japanese to English, we focus only on
translation from Japanese to English.



\citet{gordon-etal-2021-data} and \citet{Yamini-Bansal-etal-2022-ICML}
showed that the accuracy of machine translation improves as the amount
of training data or the number of model parameters increases.  What
makes patent translation different from other machine translation
domains is that numerous international patent applications are
publicly available after a certain period.  However, what we can
achieve by exploiting such resources remains unknown.

The history of creating a parallel corpus of Japanese-English patents
spans nearly 20 years.  \citet{utiyama-isahara-2007-japanese} created
a bilingual Japanese-English patent corpus of approximately 2 million
sentence pairs for the NTCIR-6 patent retrieval task
\cite{Atsushi-Fujii-etal-2007-NTCIR6}. They applied a bilingual
sentence extraction method originally developed for comparable
newspaper articles \cite{utiyama-isahara-2003-reliable} to patent
applications. These bilingual data comprised the first publicly available
large-scale Japanese-English patent corpus and were used in the NTCIR-7
patent MT task, which was the first shared task for machine
translation between Japanese and English
\cite{Atsushi-Fujii-etal-2008-NTCIR7}.

The JPO-NICT English-Japanese parallel corpus
\citeplanguageresource{JPO-NICT_parallel_corpus}, which has about 350
million Japanese-English patent sentence pairs, was jointly compiled
by the Japan Patent Office (JPO) and the National Institute of
Information and Communications Technology (NICT) from the publications
of unexamined patent applications in the United States and Japan based
on patent families.  These data, which are available to members of
Advanced Language Information Forum (ALAGIN), an organization that
resembles LDC, can be used without charge for research and development
purposes.
The JPO Patent Corpus \citeplanguageresource{JPO_Patent_Corpus} has 1M
Japanese-English patent sentence pairs and is used in the shared task of
patent translation in the Workshop on Asian Translation (WAT), which was
first held in 2015.


The JPO-NICT and JPO patent corpora were created around 2015, so they
do not reflect the latest contents and technologies. According to
\citet{utiyama-isahara-2007-japanese}, the JPO-NICT corpus includes
JPO and USPTO patents from 1993 but not after 2015. In addition, they
were made using a bilingual dictionary-based sentence alignment method
\cite{utiyama-isahara-2003-reliable}. Unfortunately, the quality of
dictionary-based alignment \cite{Daniel-Varga-etal-2005-hunalign} is
generally lower than that of translation-based alignment
\cite{sennrich-volk-2010-mt}.  State-of-the-art sentence alignment
technology could improve the quality of Japanese-English patent
copora.

We constructed JaParaPat (Japanese-English parallel patent application
corpus), which has about 350M sentence pairs from about 1.4M document
pairs from 2000 to 2021 using translation-based alignment.
International patent applications can be filed in one of two ways: the
Paris route or the PCT route.  To the best of our knowledge, ours is
the first attempt to extensively mine parallel patent applications
under both routes and align every part of the documents including
titles, abstracts, descriptions, and claims \footnote{We are releasing
  a portion of JaParaPat (covering 2016-2020) for research
  purposes at \url{https://www.kecl.ntt.co.jp/icl/lirg/japarapat/}.}  .

\section{Resources}
\begin{figure*}[!ht]
\begin{center}
\includegraphics[width=0.8\textwidth]{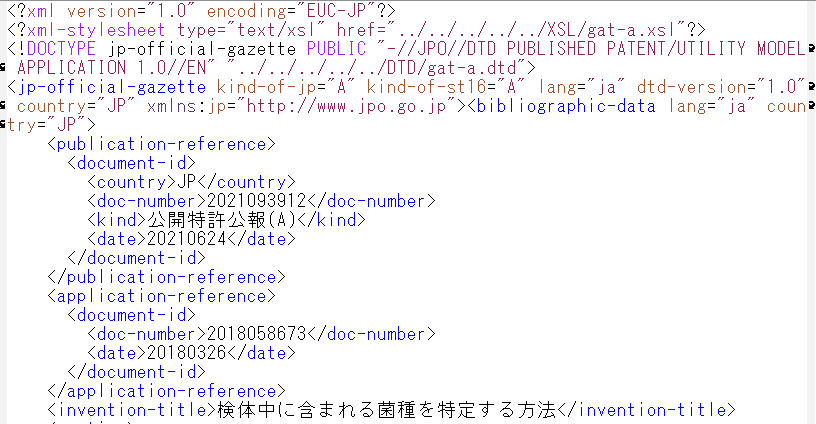} 
\includegraphics[width=0.8\textwidth]{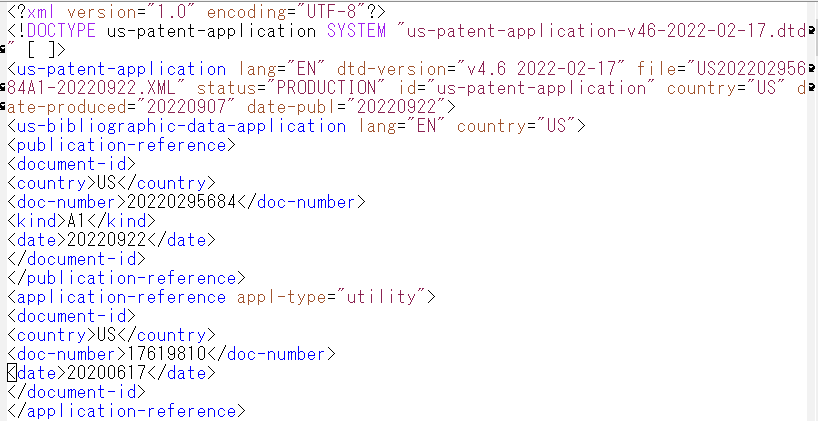} 
\caption{Example of JPO and USPTO XML files}
\label{fig:xml_ja-en}
\end{center}
\end{figure*}

\subsection{International Patent Application}

There are two ways to obtain a patent in a foreign country: directly
filing an application in that country based on the Paris Convention
(Paris route) or transferring an international application filed to a
patent office based on the Patent Cooperation Treaty (PCT route) to
that country.

Under the Paris Convention route, after filing a national application
in one country, an application is filed in another country, claiming
priority under the Paris Convention within a priority period of one
year.

In a PCT application, filing a single PCT application in a
single language using a common format to a PCT receiving office
secures priority on the filing date in every PCT member country.
However, to obtain a patent right in a country, a national phase
application must be filed within 30 months of the priority date in
that country and an examination of the patent must be undergone
following the laws of that country.  At that time, the patent
application must be translated into the language accepted by that
country's patent office.

For example, suppose a Japanese company submits a PCT application
written in Japanese to the World Intellectual Property
Organization (WIPO). In that case, JPO publishes the Japanese patent
application after entry into Japan, and USPTO publishes the English
patent application after entry into the United States.


\subsection{JPO Patent Data}

Since the Japan Patent Office (JPO) provides bulk download service of
patent information,
\footnote{\url{https://www.jpo.go.jp/system/laws/sesaku/data/download.html}}
we sent the hard drive to the patent office, which returned it with the
necessary patent information. If a company uses this system, it must
submit a company registry.\footnote{Although JPO's web page do not
  mention license conditions, we confirmed with the organization that
  we can use these data for the research and the development of
  machine translation.}

In the Japanese Patent Gazette, PCT patent applications are given a
different name than ordinary domestic applications.  A ``published
patent application'' is an ordinary domestic patent written in
Japanese.  This is the target of the Paris route searches.  A
``Japanese translation of PCT international patent application'' is a
Japanese translation of an international patent application filed with
a receiving office other than the JPO for entry into Japan.  A
``domestic re-publication of PCT international patent application'' is
an international patent application written in Japanese where JPO is
the receiving office.

On December 23, 2021, the JPO abolished the system of publishing
domestic re-publication of PCT international patent applications.
After this date, PCT applications first filed in Japan in Japanese
will only be available if they are granted as a patent after certain
amendments, so this study covers the period through 2021.

As shown in the upper part of Figure~\ref{fig:xml_ja-en}, a JPO XML
file represents each patent data by jp-official-gazette
element.\footnote{\url{https://www.jpo.go.jp/system/laws/koho/shiyo/kouhou_siyou_vol4-7.html}}
The kind-of-jp attribute is the gazette type.  A is a published patent
application, T is a Japanese translation of PCT international patent
application, and S is a domestic re-publication of PCT international
patent application.

Bibliographic information is found in the bibliographic-data element.
For documents whose kind-of-jp attribute is A or T, the publication
number is obtained from the publication-reference element and the
application number is obtained from the application-reference element.
For documents whose kind-of-jp attribute is T, the application number
is obtained from the pct-or-regional-filing-data element and the
publication number is obtained from the
pct-or-regional-publishing-data element.

We extracted the text enclosed by the p tags of the XML elements
corresponding to the patent's title, abstract, description, and
claim. In other words, for sentence alignment, we excluded the claim
numbers, the paragraph numbers, the mathematical expressions, figures,
the etc. Since January 2004, Japanese patent applications have been
filed in the XML format.  Before 2004, they were in the SGML
format. We veryfied that data in the SGML format have the same
extraction targets as in the XML format.


\subsection{USPTO Patent Data}

The United States Patent and Trademark Office (USPTO) provides patent
application full text data.
\footnote{\url{https://developer.uspto.gov/product/patent-application-full-text-dataxml}}
We can obtain the documentation and the DTD from USPTO's web
page.\footnote{\url{https://www.uspto.gov/learning-and-resources/xml-resources}}
USPTO provides patent application full text data from March 15, 2001.
Since corresponding patent applications may have been published in
Japan one year before they were published in the U.S., this study
covers the period from 2000.

As shown in the lower part of Figure~\ref{fig:xml_ja-en}, a USPTO XML
file represents each patent by us-patent-application element.
Bibliographic information is in the us-bibliographic-data-application
element.  The application number is obtained from the
application-reference element, and the publication number is obtained
from the publication-reference element.

If a pct-or-regional-filing-data element exists and its doc-number
attribute begins with PCT, such as "PCT/JP2005/003817," we consider it
a PCT application, and the value of the doc-number attribute is its
application number.  The USPTO's PCT patent application does not have
the same distinction as that between T and S in the JPO's kind-of-jp
attribute.


\subsection{EPO DOCDB}
\begin{figure}[!ht]
\begin{center}
\includegraphics[width=0.8\columnwidth]{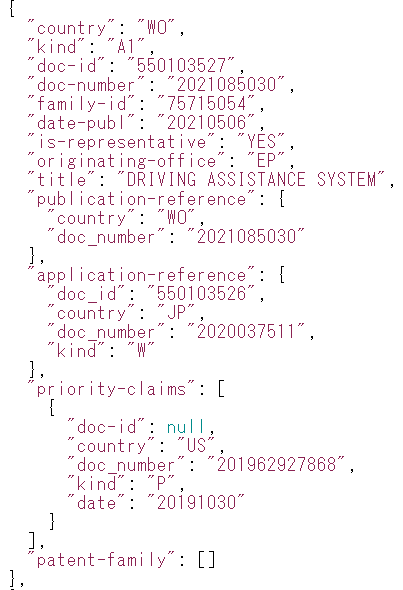} 
\caption{Example of information extracted from exch:exchange-document
  element in DOCDB}
\label{fig:epo_docdb}
\end{center}
\end{figure}

The European Patent Office (EPO) provides (for a fee) worldwide bibliographic data of
patents (DOCDB). We can obtain a sample of
DOCDB\footnote{\url{https://www.epo.org/searching-for-patents/data/bulk-data-sets/docdb.html}}
and its manual
\footnote{\url{https://www.epo.org/searching-for-patents/data/bulk-data-sets/manuals.html}}
from its web site.


We obtained DOCDB, as of April 2022, to get information on patent
families.  A patent family is a set of patents obtained in various
countries to protect a single invention.  We obtained a patent
family by analyzing the priority claim data in the DOCDB.

A DOCDB XML file represents each patent by exch:exchange-document
element.  The priority-claims information is aggregated in the
exch:priority-claims element under the exch:bibliographic-data
element.  Figure~\ref{fig:epo_docdb} is an example of information
extracted from exch:exchange-document element in a DOCDB XML file.  We
extracted the country, doc-number, kind, and date attributes of the
document-id element, which is the subject of the priority claim.
Kind-code A is an ordinary patent application, and W is a PCT
application.


\section{Methodology}

\subsection{Document Alignment}
\label{sec:document_alignment}
We mapped the patent applications published by the JPO and USPTO based
on the patent families obtained from the EPO's DOCDB.  The original data are
all in XML, and we implemented the document alignment procedure
described below using the xml.etree.ElementTree module in the python
standard library.

We considered pairs of Japanese and English patent applications in the
same patent family to be translations of each other.  If there are
more than one such pairs, we selected the oldest document pair because
a set of documents claiming priority for the same document is almost
always a modified version of the initial application.

The search method for a bilingual document pair differs slightly
between the Paris route and the PCT routes.
The primary example of the Paris route is where one application claims
priority based on another.  A US patent that claims priority based on
one filed in Japan is a patent in DOCDB where the country
attribute of the exchange-document element is US and the country
attribute and kind attribute in the priority-claims element are JP and
A, respectively.  The same is true for a Japanese patent that claims
priority based on one filed in the US. In this paper, we refer to
the former as 'jp-us' and the latter as 'us-jp' based on the order in
which the patents were filed in the countries.

We extracted a pair of Japanese and U.S. patent applications that
claims priority based on a shared third patent application, such as a
patent that is first filed in China and then filed in Japan and the
U.S.  For these cases, we first listed a pair of the document-id in the
exchange-document element and the document-id in the priority-claim element, for
all Japanese and U.S. patent applications.  We then extracted JP-US
patent application pairs with the same document-id in the priority-claim
element. In this paper, we refer to such pairs as 'jp-x-us' where x
indicates that a shared third patent application exists.

For the PCT route, we first extracted from the DOCDB applications
where the kind attribute of the application-reference element is W.
We extracted applications from the JPO where the kind-code attribute
is S or T and the doc-number starts with WO.  We extracted
applications from the USPTO where the pct-or-regional-filing-data
starts with PCT.  If the application number obtained from the JPO data
and the application number obtained from the USPTO data are the same
and exist in the DOCDB, we consider the Japanese and the U.S. patent
applications to be translations of each other. In this paper, we refer
to all PCT applications as 'pct'.


\subsection{Sentence Alignment}

We used two methods for sentence alignment: one based on bilingual
dictionaries \cite{utiyama-isahara-2003-reliable} and another based on
machine translation \cite{sennrich-volk-2010-mt}.  We first obtained a
bilingual patent data using a dictionary-based sentence alignment
method and trained a translation model from the bilingual patent data
and JParaCrawl \cite{morishita-etal-2022-jparacrawl}, a publicly
available large-scale Japanese-English parallel corpus collected from
the web.  We then obtained the final bilingual patent data using a
translation-based sentence alignment method.


We divided the Japanese and U.S. patent applications into titles, abstracts,
descriptions, and claims and aligned them separately. We used
split-sentences.perl in Moses for sentence segmentation in both
Japanese and English.
\footnote{\url{https://github.com/moses-smt/mosesdecoder/blob/master/scripts/ems/support/split-sentences.perl}}

For our dictionary-based sentence alignment, we used our
implementation of \citet{utiyama-isahara-2003-reliable}'s method. As
for bilingual dictionary, we used a Japanese-English dictionary of EDR
with 1,690,174 entries \citeplanguageresource{EDR}.  We used
mecab-unidic\footnote{\url{https://taku910.github.io/mecab/}} for
Japanese word segmentation and
TreeTagger\footnote{\url{https://www.cis.uni-muenchen.de/~schmid/tools/TreeTagger/}}
for English tokenization.

For translation-based sentence alignment, we used
Bleualign\footnote{\url{https://github.com/rsennrich/Bleualign}}.
We used fairseq \cite{ott-etal-2019-fairseq} for machine translation.

\section{JaParaPat Overview}

\subsection{Data Statistics}
\begin{table*}[!ht]
  \centering
  \begin{tabular}{r|rrrr|rrrr}
    \hline
    & \multicolumn{4}{c|}{Sentence pairs} & \multicolumn{4}{c}{Document pairs} \\
    & jp-us & jp-x-us & us-jp & pct & jp-us & jp-x-us & us-jp & pct \\
\hline
2000 & 804,586 & 116,806 & & 92,242 & 4,189 & 865 & & 402 \\
2001 & 1,936,229 & 423,355 & 842,701 & 122,205 & 11,223 & 3,249 & 5,608 & 550 \\
2002 & 2,599,128 & 1,161,071 & 3,181,974 & 51,214 & 14,385 & 8,521 & 18,941 & 200 \\
2003 & 2,216,059 & 1,944,235 & 4,083,604 & 1,975,669 & 11,755 & 12,506 & 22,385 & 7,743 \\
2004 & 2,719,911 & 860,287 & 3,848,196 & 4,319,575 & 16,126 & 7,542 & 23,324 & 18,978 \\
2005 & 2,352,235 & 994,049 & 5,024,330 & 4,977,803 & 12,973 & 8,193 & 28,089 & 20,647 \\
2006 & 2,297,878 & 1,131,340 & 5,770,905 & 4,513,947 & 12,239 & 8,810 & 30,832 & 18,469 \\
2007 & 2,513,900 & 1,081,103 & 5,883,197 & 5,050,197 & 13,124 & 8,147 & 30,481 & 20,444 \\
2008 & 2,535,483 & 921,678 & 5,752,965 & 8,264,349 & 12,956 & 6,715 & 29,165 & 31,506 \\
2009 & 1,813,767 & 861,456 & 6,259,067 & 8,227,809 & 9,180 & 6,049 & 31,303 & 31,304 \\
2010 & 1,559,327 & 821,388 & 6,310,667 & 8,178,496 & 7,381 & 5,169 & 29,025 & 29,196 \\
2011 & 1,869,428 & 957,781 & 6,739,639 & 6,497,215 & 8,341 & 5,789 & 28,899 & 22,932 \\
2012 & 1,990,833 & 945,927 & 7,252,931 & 7,781,432 & 8,868 & 5,560 & 30,065 & 27,381 \\
2013 & 2,363,076 & 1,012,462 & 6,598,196 & 10,278,504 & 10,050 & 6,021 & 28,101 & 35,850 \\
2014 & 2,144,452 & 1,116,288 & 6,651,888 & 8,055,146 & 9,168 & 6,088 & 26,716 & 27,326 \\
2015 & 2,506,286 & 1,030,098 & 6,754,694 & 9,391,589 & 10,314 & 5,229 & 26,087 & 31,380 \\
2016 & 2,494,488 & 1,017,181 & 5,746,295 & 9,313,031 & 10,233 & 4,988 & 22,317 & 29,196 \\
2017 & 4,861,052 & 1,017,358 & 3,624,756 & 16,251,900 & 19,876 & 5,045 & 14,467 & 51,791 \\
2018 & 3,284,674 & 918,138 & 5,153,238 & 11,696,010 & 12,625 & 4,369 & 19,239 & 35,822 \\
2019 & 3,227,271 & 1,066,833 & 6,107,334 & 12,483,342 & 12,388 & 5,251 & 23,685 & 36,961 \\
2020 & 3,740,996 & 1,093,506 & 4,251,027 & 11,962,022 & 13,306 & 4,781 & 15,032 & 34,006 \\
2021 & 1,043,944 & 849,489 & 4,838,957 & 11,275,167 & 3,656 & 3,818 & 16,928 & 30,884 \\
\hline
sum & 52,875,003 & 21,341,829 & 110,676,561 & 160,758,864 & 244,356 & 132,705 & 500,689 & 542,968 \\
    & \multicolumn{4}{c|}{345,652,257} & \multicolumn{4}{c}{1,420,718} \\
\hline
  \end{tabular}
  \caption{Number of parallel sentence and document pairs collected
    annually from 2000 to 2021}
  \label{tab:data_statistics}
\end{table*}

Table~\ref{tab:data_statistics} shows the number of annually collected
document and sentence pairs from 2000 to 2001. In this table, the
numbers are divided into jp-us, jp-x-us, us-jp, and pct, as described
in Section~\ref{sec:document_alignment}.  Here, the years are based on
the publication year of the Japanese patent applications.

The parallel corpus has about 350M sentence pairs from about 1.4M
document pairs.  Since the USPTO U.S. patent data are only available
after 2001, no Japanese patent applicationss published in 2000 have
any available U.S. patent applications as priority claims.  Since we
used the DOCDB as of April 2022, the patent family are incomplete on
the applications published in Japan in 2021.  Thus, scant parallel
data exist for 2021.

The ratio of Paris routes to PCT routes in the parallel corpus is
almost one-to-one. The former route has more document pais, but the
latter route has more sentence pairs because document pairs in the
Paris route are not necessarily translations of each other, while
document pairs in the PCT route must be translations of each
other. In general, we extracted 60-70\% of the sentences as parallel
sentence pairs from the Japanese and English document pairs.
Within the Paris route, The amount of bilingual data for us-jp
is the largest, followed by jp-us and jp-x-us.

\subsection{Data Format}
\begin{figure*}[!ht]
\begin{center}
\includegraphics[width=\textwidth]{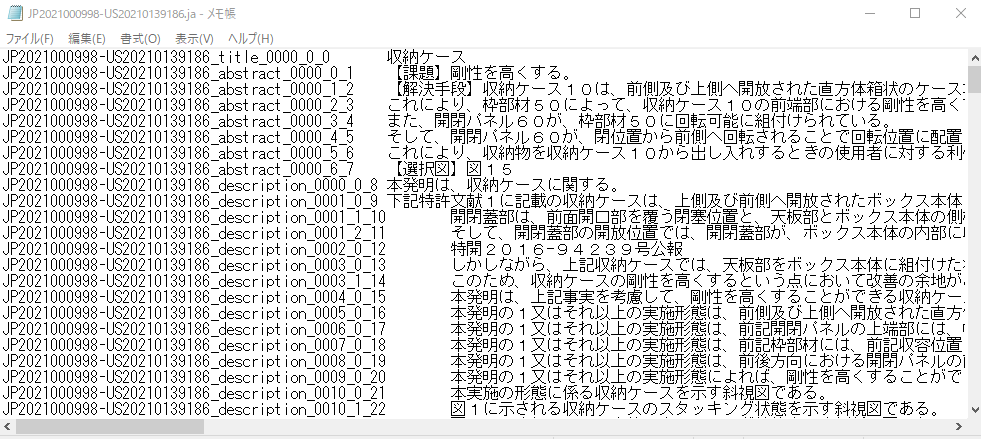} 
\includegraphics[width=\textwidth]{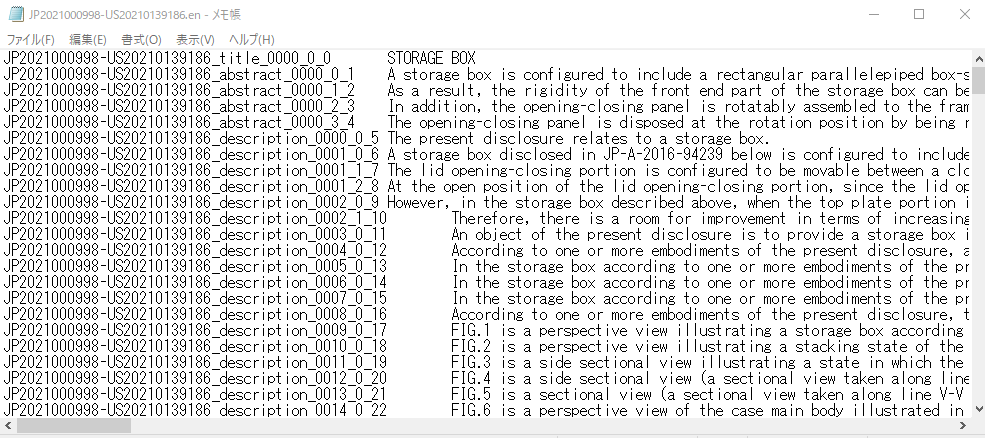} 
\caption{Example of Japanese and English text files for a patent
  document pair}
\label{fig:text_ja-en}
\end{center}
\end{figure*}
\begin{figure*}[!ht]
\begin{center}
\includegraphics[width=\textwidth]{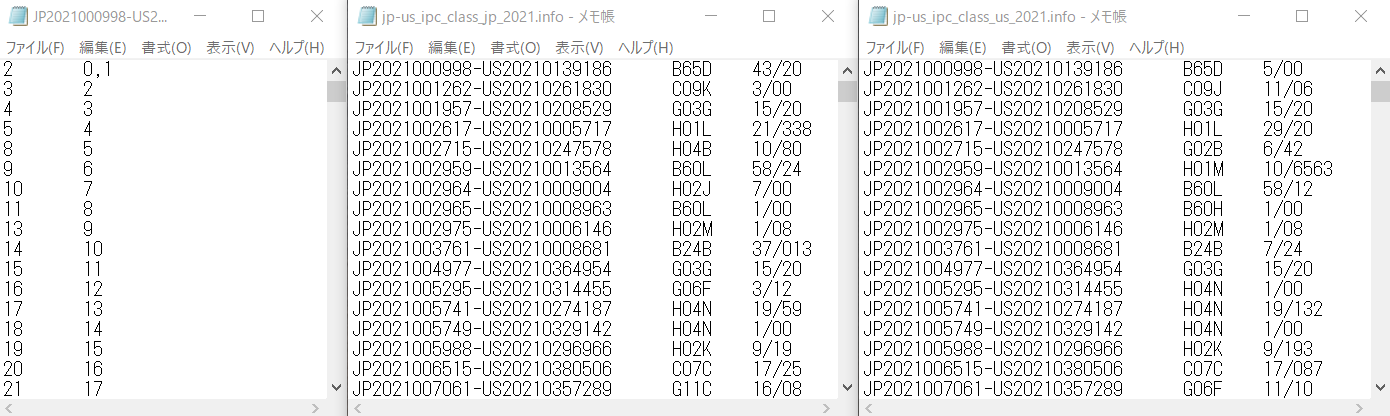} 
\caption{Example of sentence alignment file for a patent document pair
and IPC data for patent document pairs}
\label{fig:align_ja-en}
\end{center}
\end{figure*}

Figure~\ref{fig:text_ja-en} shows an example of Japanese and English
text files for a patent document pair.  We first assigned a pair of
publication numbers in Japan and the U.S. as an ID for a parallel
document pair, such as JP2021000998-US20210139186.  We divided Japanese
and U.S. patent documents into four parts: title, abstract,
description, and claim, separated each part into paragraphs and
sentences, and finally assigned a concatenation of a document pair ID,
a part, a paragraph number, a sentence number within a paragraph, and
a sentence number within a document as an ID to a sentence.

The leftmost screenshot in Figure~\ref{fig:align_ja-en} shows an example
of a sentence alignment file for a patent document pair.  The first
column represents the sentence number within a Japanese document, and
the second column represents the sentence number within an English
document.  Multiple numbers in one column represent a many-to-many
alignment.  This configuration allows us to create a claim-specific
translation model by extracting only the sentence pairs in the claim,
or a context-aware translation model by extracting consecutive
sentence pairs in the same paragraph.

The middle and rightmost screenshots in Figure~\ref{fig:align_ja-en}
shows examples of International Patent Classification (IPC) data for
each document pairs.  This information allows us to create a
translation model dedicated to a specific field.

\section{Experiments}
\subsection{Training and Test Data}

\begin{table*}[htb]
\centering
\begin{tabular}{lrrr}
  \hline
  route	& documents 	& sentences	& words \\
  \hline
  Paris & 866,931 & 181,907,843 & 7,378,214,793 \\
  PCT   & 527,068 & 154,860,596 & 6,180,045,629\\
  Paris+PCT & 1,393,999 & 336,768,439 & 13,558,260,422 \\
\hline
\end{tabular}
\caption{Number of document pairs, sentence pairs, and words on
  English side in the training data}
\label{tab:training_data}
\end{table*}

\begin{table}
\centering
\begin{tabular}{lrrll}
  \hline
  Test data	& \#sentences	& \#words \\
  \hline
  Paris SH2021  & 1,000 & 37,990 \\
  PCT SH2021    & 1,000 & 38,676 \\
  In-house test1& 1,002	& 33,405 \\
  In-house test2&   988	& 26,945 \\
  ASPEC test	& 1,812	& 39,573\\
  \hline
\end{tabular}
\caption{Number of sentences and words on English side in the test sets}
\label{tab:test_data}
\end{table}

To confirm the quality of JaParaPat, we conducted translation
experiments from Japanese to English.  Table~\ref{tab:training_data}
shows the number of document pairs, sentence pairs, and the number of
words on the English side of the training data for the translation
model.  We used the sentence pairs from 2000 to the first half of 2021
to train the translation models.

Table~\ref{tab:test_data} shows the number of sentences and words on
the English side of the test data. We randomly sampled 1,000 sentences
for the test data and 2,000 sentences for the validation data from the
second half of 2021 in the Paris and PCT routes, respectively.  Note
that while these Paris and PCT test sets cover a wide range of topics,
they are not guaranteed to be parallel sentence pairs because they are
automatically extracted and sampled.

We also used as test data the in-house Japanese PCT patent
applications published or to be published in 2022 or later and their
translations into English by two translation companies specializing in
patent translation.  The target domain is information and
communication technology (ICT) and includes a wide range of content
from hardware to software.  Preliminary studies revealed that the
scores of automated evaluations varied by translation companies, not
by content, so we created a test set for each translation company.

We also used test sentences from the Asian Scientific Paper Excerpt
Corpus (ASPEC) \cite{nakazawa-etal-2016-aspec} as publicly available
out-of-domain test data.  There are no publicly available in-domain
(patent) test data suitable for the quality assessment of our parallel
corpus.  Since our training data covers from 2000 to the first half of
2021, the test data should be Japanese patent applications published
in the second half of 2021 or later.  However, the JPO Patent Corpus
test set used in the patent translation shared task of WAT-2023 was
made from patent documents published in 2019-2020, which was likely to
be included in our training data.

\subsection{Training Conditions}
\begin{table}[htb]
  \small
  \centering
  \begin{tabular}{ll}
    \hline
    architecture	& transformer\_wmt\_en\_de\_big \\
    enc-dec layers	& 6 \\
    optimizer		& Adam ($\beta_{1}=0.9, \beta_{2}=0.98$) \\
    learning rate schedule	& inverse square root decay \\
    warmup steps	& 4,000 \\
    max learning rate	& 0.001 \\
    dropout		& 0.3 \\
    gradient clip	& 0.1 \\
    batch size		& 1M tokens \\
    max number of updates	& 60K steps \\
    validate interval updates	& 1K steps \\
    patience		& 5 \\
    \hline
  \end{tabular}
  \caption{List of hyperparameters for the Transformer}
  \label{tab:parameters}
\end{table}

We used fairseq \cite{ott-etal-2019-fairseq} for machine translation.
The translation model is Transformer big
\cite{vaswani-etal-2017-transformer}. Table\ref{tab:parameters} shows
the hyperparameters of the Transformer. The translation models in this
paper were all trained under this condition.  We used sentencepiece
\cite{kudo-richardson-2018-sentencepiece} for tokenization.  We
randomly sampled 7M sentence pairs from the patent corpus and 3M
sentence pairs from JParaCrawl to train the sentencepiece model. The
vocabulary size was 32K for both Japanese and English. We set the
character\_coverage to 0.9995 and the byte\_fallback to true.  We used
both sacreBLEU \cite{papineni-etal-2002-bleu,post-2018-call} and COMET
\cite{rei-etal-2020-comet} for evaluation, but we mainly used BLEU
because choosing the appropriate technical terms is essential in
patent translation.

\subsection{Comparison of Sentence Alignment Methods}

\begin{table*}
\centering
\begin{tabular}{lrrrr}
  \hline
  training data				& test1	& test2	& pairs	& updates \\
  \hline
  2000-2013Paris\_dict			& 62.6	& 51.5	& 34M	& 17K \\
  2000-2013Paris\_dict+JParaCrawl	& 63.6	& 54.0	& 56M	& 26K \\
  2000-2013Paris\_trans			& 63.4	& 53.0	& 43M	& 16K \\
  \hline
\end{tabular}
\caption{Comparison of Sentence Alignment Methods}
\label{tab:initial_model}
\end{table*}

First, we examined the accuracy of the translation model used in our
translation-based sentence alignment method.  We collected about 34M
sentence pairs (2000-2013Paris\_dict) from document pairs in the Paris
route from 2000 to 2013 using a dictionary-based sentence alignment
method. We then created a translation model trained on these 34M
patent sentence pairs and JParaCrawl
(2000-2013Paris\_dict+JParaCrawl).  Using this translation model for
translation-based sentence alignment, we collected about 43M sentence
pairs (2000-2013Paris\_trans) from the same document pairs used for
dictionary-based sentence alignment.

Table\ref{tab:initial_model} shows that the translation accuracy
(BLEU) improved when we combined the sentence pairs from the patent
applications and the web. When we use translation-based sentence
alignment, we collected more sentence pairs (34M to 43M) with higher
quality (62.6/51.5 to 63.4/53.0) than dictionary-based sentence
alignment. Recent research shows that translation-based sentence
alignment method can obtain better and more bilingual sentence pairs
than dictionary-based method
\cite{banon-etal-2020-paracrawl,morishita-etal-2022-jparacrawl} and we
confirmed this finding in our experiment.


Test1 and test2 differed in sacreBLEU by 10 points in the models
trained on the patent corpus.  Since both translation companies
manually post-edited the output of their patent translation systems,
we assume that the differences in the machine translation and
post-editing methods significantly impacted the automatic evaluation
measurements. The results indicate that post-editing bias may be a
problem in the future for parallel corpora collected from patent
applications because more and more patent translation companies are
adopting machine translation post-editing.

\subsection{Japanese-to-English Translation Accuracy}
\begin{table*}[htb]
  \small
  \centering
  \begin{tabular}{lrrrrrrrrrrll}
    \hline
    training data
    & \multicolumn{2}{c}{Paris} & \multicolumn{2}{c}{PCT}
    & \multicolumn{2}{c}{test1} & \multicolumn{2}{c}{test2}
    & \multicolumn{2}{c}{ASPEC} & pairs & updates \\
    & bleu & comet & bleu & comet
    & bleu & comet & bleu & comet
           & bleu & comet & & \\
    \hline
    JParaCrawl(JPC)
    & 31.9 & 0.817 & 35.6 & 0.827
    & 36.2 & 0.838 & 35.8 & 0.826
           & 20.6 & \textbf{0.828}
                          & 22M & 20K \\
    
    Paris
    & \textbf{55.6} & \textbf{0.867} & 56.5 & \textbf{0.877}
    & 66.8 & \textbf{0.881} & 53.2 & 0.820
           & 20.5 & 0.823
                          & 182M & 44K \\
    PCT
    & 52.7 & 0.857 & \textbf{57.3} & 0.873
    & 64.6 & 0.866 & 51.6 & 0.811
           & 20.6 & 0.820
                          & 155M & 53K \\
    Paris+PCT
    & 55.5 & 0.864 & 55.7 & 0.872
    & 67.0 & 0.876 & 46.0 & 0.820
           & 20.8 & 0.821
                          & 337M & 57K \\
    JPC+Paris+PCT
    & 54.7 & 0.863 & 56.0 & 0.872
    & \textbf{67.7} & 0.880 & \textbf{55.5} & \textbf{0.846}
           & \textbf{21.3} & 0.827
                          & 359M & 42K \\
    \hline
  \end{tabular}
  \caption{Comparison of translation accuracies with respec to the size
    of training data}
  \label{tab:accuracies}
\end{table*}

Table~\ref{tab:accuracies} shows the translation accuracy of the model
trained from the collected patent sentence pairs.  Compared to
JParaCrawl, JaParaPat improved the patent translation accuracy by 20
bleu points.  Comparing the Paris route and the PCT routes, although
the amount of data is almost identical (around 150M), the Paris route
has generally higher translation accuracy.  We assume this result is
because the Paris route contains a greater variety of patent
applications since the PCT route is mainly used by large companies.

Training the translation model from more than 300M patent bilinguals
from both the Paris and PCT routes improved translation accuracy,
although the improvement is moderate and unstable.  However, when we
added 22M web-crawled sentence pairs of JParaCrawl to 337M patent
sentence pairs of JaParaPat, the translation accuracy of test2 and
ASPEC increased, suggesting that the patent sentence pairs lack
diversity. We observed that the perplexity of the patent texts is low
compared to that of web texts. Adding web text makes the patent
translation model more robust than increasing the amount of patent
text.


\section{Related Works} 


\subsection{Patent Parallel Corpus}

With the increasing popularity of the PCT international patent
applications and such new technologies as sentence alignment using
neural machine translation models, a different approach has recently
emerged for creating a parallel patent corpus. In 2011, World
Intellectual Property Organization (WIPO) created the Corpus Of
Parallel Patent Applications (COPPA) from the titles and abstracts of
PCT applications. COPPA V2.0 \cite{Junczyz-Dowmunt-etal-2016-COPPA}
consists of eight language pairs, mainly English, and has about 1
million sentence pairs of Japanese-English data.  ParaPat
\cite{soares-etal-2020-parapat} is a bilingual data set of 22 language
pairs created from patent abstracts in Google Patents, with 17M
sentence pairs of Japanese-English data.  COPPA and ParaPat use
Hunalign \cite{Daniel-Varga-etal-2005-hunalign}, a dictionary-based
sentence alignment tool.

EuroPat \cite{heafield-etal-2022-europat} is a parallel patent corpus
of six European language as well as English collected from USPTO and
EPO.  It extracts sentence pairs from granted patents with an emphasis
on quality.  It uses the API provided by the EPO to obtain patent
families for document alignment and translates non-English documents
into English for sentence alignment with
Bleualign-cpp\footnote{\url{https://github.com/bitextor/bleualign-cpp}},
a translation-based sentence alignment tool developed in the ParaCrawl
Project \cite{banon-etal-2020-paracrawl}.

Our approach resembles EuroPat, although we used unexamined
patent applications rather than granted patents and made alignments
between Japanese and English rather than among European
languages. 

\subsection{Japanese-English Parallel Corpus}

In areas other than patents, ASPEC \cite{nakazawa-etal-2016-aspec} is
one of the first publicly available Japanese-English parallel
corpora. It is comprised of English summaries attached to Japanese
scientific and technical papers.  Its domain is close to patents, but
it only has 3 million sentence pairs.  ASPEC has been used in a shared
task of WAT since 2014 \cite{nakazawa-etal-2014-overview}.

JParaCrawl \cite{morishita-etal-2022-jparacrawl} is one of the largest
publicly available Japanese-English parallel corpora.  It is a
web-crawled corpus that contains a wide variety of domains.
JParaCrawl has been used in news translation task and general machine
translation task in WMT since 2020 \cite{barrault-etal-2020-findings}.

Although the JPO-NICT corpus is one of the largest publicly available
Japanese-English parallel patent corpora, its construction is unknown
since it has not been published as a technical paper.  Assuming that
this corpus was made from a procedure similar to the NICIR-7 PATMT
\cite{utiyama-isahara-2007-japanese}, it identifies Japanese patents
by the priority number listed in the U.S. patents. Thus, this corpus
only covers the jp-us of the Paris route in our term. It used
dictionary-based sentence alignment, while we used sentence-based
alignment.


The newly created JaParaPat is one of the largest and highest-quality
Japanese-English patent parallel corpora.  It will serve as the
foundation for future machine translation research in the science and
technology field.

\subsection{Sentence Alignment}

Sentence alignment can be classified into three categories: a
bilingual dictionary-based method
\cite{utiyama-isahara-2003-reliable,Daniel-Varga-etal-2005-hunalign}
such as hunalign, a machine translation-based method
\cite{sennrich-volk-2010-mt} such as Bleualign, or a multilingual
sentence embedding-based method
\cite{thompson-koehn-2019-vecalign,chousa-etal-2020-spanalign} such as
Vecalign.

Although the sentence embedding-based method is the most accurate
approach, it is unfortunately also the most computationally expensive.
Since we must process a large amount of data in this work, we used a
translation-based method to balance speed and accuracy.

\section{Conclusion}
We extracted patent sentence pairs as exhaustively as possible from
Japanese and U.S. patent applications from 2000-2021 and constructed a
parallel patent corpus of more than 300M sentence pairs.

By training a translation model on the parallel patent corpus, we
improved the patent translation accuracy by about 20 bleu points
compared to JParaCrawl by using 22M sentence pairs collected from the
web. We collected more and better sentence pairs by using a
translation-based sentence alignment method compared to a
dictionary-based sentence alignment method.

Future work includes increasing the number of parameters in the
translation model and designing a filter to remove noise in the
parallel corpus to improve translation accuracy with reference to the
study of data scaling laws
\cite{gordon-etal-2021-data,Yamini-Bansal-etal-2022-ICML}.

\clearpage
\nocite{*}
\section{Bibliographical References}\label{sec:reference}

\bibliographystyle{lrec-coling2024-natbib}
\bibliography{240326slim-anthology,240326custom}

\section{Language Resource References}
\label{lr:ref}
\bibliographystylelanguageresource{lrec-coling2024-natbib}
\bibliographylanguageresource{languageresource}

\clearpage
\appendix
\section{Statistics of the Public Version}
\begin{table}[!ht]
  \centering
  \begin{tabular}{r|rrrr|r}
    \hline
    & jp-us & jp-x-us & us-jp & pct & sum \\
\hline
2016 & 7,241,502 & 1,322,124 & 1,181,150 & 10,287,313 & 20,032,089 \\
2017 & 7,892,204 & 1,399,012 & 1,226,177 & 10,354,135 & 20,871,528 \\
2018 & 7,639,692 & 1,262,972 & 1,044,728 & 11,171,128 & 21,118,520 \\
2019 & 8,867,148 & 1,450,851 & 1,157,361 & 11,625,720 & 23,101,080 \\
2020 & 8,617,540 & 1,570,684 & 1,088,832 & 10,843,470 & 22,120,526 \\
\hline
sum  & 40,258,086 & 7,005,643 & 5,698,248 & 54,281,766 & 107,243,743 \\
  \end{tabular}
  \caption{Number of sentence pairs}
  \label{tab:public_version}
\end{table}
Table~\ref{tab:public_version} presents the number of sentence pairs
available in the public version of JaParaPat. In the public version,
we modified the document alignment, sentence segmentation, and
sentence alignment of patent applications, leading to a discrepancy in
the number of sentence pairs compared to
Table~\ref{tab:data_statistics}.






\end{document}